\def\year{2022}\relax
\documentclass[letterpaper]{article} 
\usepackage{aaai22}  
\usepackage{times}  
\usepackage{helvet}  
\usepackage{courier}  
\usepackage[hyphens]{url}  
\usepackage{graphicx} 
\usepackage{natbib}  
\usepackage{caption} 
\DeclareCaptionStyle{ruled}{labelfont=normalfont,labelsep=colon,strut=off} 
\usepackage{color}
\usepackage[table,xcdraw]{xcolor}
\usepackage{algorithm}
\usepackage{algorithmic}
\usepackage{multirow}

\usepackage{bm}
\usepackage{amsmath}

\usepackage[colorlinks,
            linkcolor=red,       
            anchorcolor=red,  
            citecolor=blue,        
            ]{hyperref}
\usepackage{newfloat}
\usepackage{listings}
\floatstyle{ruled}
\newfloat{listing}{tb}{lst}{}
\floatname{listing}{Listing}
\title{ProgressiveMotionSeg: Mutually Reinforced Framework for Event-Based Motion Segmentation}
\author {
    Jinze Chen\textsuperscript{\rm 1}\equalcontrib,
    Yang Wang\textsuperscript{\rm 1}\equalcontrib,
    Yang Cao\textsuperscript{\rm 1,2\textdagger},
    Feng Wu\textsuperscript{\rm 1},
    Zheng-Jun Zha\textsuperscript{\rm 1}\thanks{Corresponding author}
}
\affiliations {
    \textsuperscript{\rm 1}University of Science and Technology of China,\\
    \textsuperscript{\rm 2}Institute of Artificial Intelligence, Hefei Comprehensive National Science Center\\
    chjz@mail.ustc.edu.cn, \{ywang120,forrest,fengwu,zhazj\}@ustc.edu.cn
}

\usepackage{bibentry}

\newcommand{\R}[1]{\textcolor{red}{#1}}
\renewcommand{\R}[1]{#1}

\begin{document}

\maketitle

\begin{abstract}
Dynamic Vision Sensor (DVS) can asynchronously output the events reflecting apparent motion of objects with microsecond resolution, and shows great application potential in monitoring and other fields. However, the output event stream of existing DVS inevitably contains background activity noise (BA noise) due to dark current and junction leakage current, which will affect the temporal correlation of objects, resulting in deteriorated motion estimation performance. Particularly, the existing filter-based denoising methods cannot be directly applied to suppress the noise in event stream, since there is no spatial correlation. To address this issue, this paper presents a novel progressive framework, in which a Motion Estimation (ME) module and an Event Denoising (ED) module are jointly optimized in a mutually reinforced manner. Specifically, based on the maximum sharpness criterion, ME module divides the input event into several segments by adaptive clustering in a motion compensating warp field, and captures the temporal correlation of event stream according to the clustered motion parameters. Taking temporal correlation as guidance, ED module calculates the confidence that each event belongs to real activity events, and transmits it to ME module to update energy function of motion segmentation for noise suppression. The two steps are iteratively updated until stable motion segmentation results are obtained. Extensive experimental results on both synthetic and real datasets demonstrate the superiority of our proposed approaches against the State-Of-The-Art (SOTA) methods.
\end{abstract}

\begin{figure}
    \centering
    \includegraphics[width=0.40\textwidth]{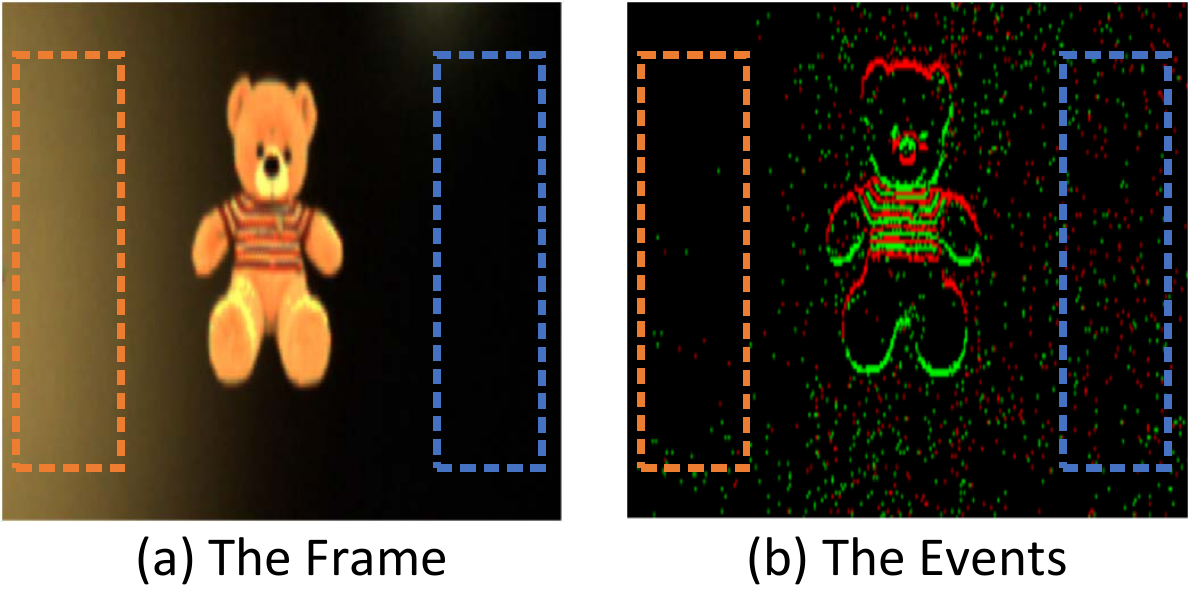}
    \caption{The noise is unavoidable in the output of DVS, as shown in the orange and blue box area. Besides, the noise level is affected by the brightness level, the lower the brightness, the greater the noise level, as shown in the blue box.}
    \label{fig:ba_noise}
\end{figure}
\section{Introduction}
The Dynamic Vision Sensors (DVS) are novel bio-inspired vision sensors that record the intensity changes asynchronously in microsecond resolution and output a series of positive or negative binary spikes (called events) representing the relative motion cues between the camera and objects \cite{tpami2020survey}. Benefited to its high-sensitivity to intensity changes, DVS is well suited for conducting motion segmentation that requires high temporal resolution, \emph{e.g.}, high-speed aerial vehicles tracking and detection \cite{kim2016real, vidal2018ultimate}. 

Given the events as input, there are mainly two strategies to perform motion segmentation. One way transforms the events into intensity frames and then conducts the motion segmentation on them as in traditional frame-based cameras \cite{kepple2020jointly, mitrokhin2019ev}. Another feasible strategy directly performs the motion segmentation in event space, which requires firstly clustering all the events into diﬀerent groups and then calculating the respective motions \cite{stoffregen2019event}. However, none of these methods consider the effect of Background Activity (BA) noise produced by dark current and junction leakage current even if there is no change in light intensity \cite{nozaki2017temperature}. The BA noise appears sparser and randomly distributed, as shown in Fig.\ref{fig:ba_noise}, which will destroy the temporal correlation of real activity events, resulting in the motion segmentation performance degradation. Worse still, the BA noise level is coupled with the brightness level due to the logarithmic compression in the front-end photoreceptor \cite{patrick2008128x}. The lower the brightness level is, the higher the BA noise level is, as shown in the orange and blue box area of Fig.\ref{fig:ba_noise} (b), which will further increase the difficulty of motion segmentation.

In practice, one straightforward solution for this problem is to first filter out BA noise using denoising methods \cite{khodamoradi2018n, feng2020event, wang2020joint} and then perform motion segmentation on remaining events. However, the events lack spatial correlation due to the sparsity, so we cannot directly use traditional image denoising methods for event denoising. Besides, the specially designed event denoising methods mainly utilize the spatial-temporal correlation on local regions to perform event denoising, which cannot capture long-time temporal correlation, which is essential for motion segmentation. 

To address this problem, this paper proposes a novel progressive framework in which a Motion Estimation (ME) module and an Event Denoising (ED) module are jointly optimized to improve the motion segmentation performance. Specifically, the ME module first performs adaptive event clustering in a motion compensating warp field under the constraint of maximum sharpness criterion and then outputs the temporal correlation by exploring the clustered motion parameters. The temporal correlation is transmitted to ED module and serves as guidance to help ED module to perceive the noise distribution and output it as confidence maps, in which the greater the confidence value, the more likely the event is a real activity event. The confidence is multiplied to the event streams, and re-weighted event streams are transmitted to the ME module to suppress noise's influence on motion estimation. The above two processes are progressively performed until obtaining stable results, and each of the preceding processes benefits from the gradually improved results in the other. Experimental results on both synthetic and real event sequences demonstrate the superiority of our proposed method against the SOTA methods. The contributions of this work are three-fold: 

\begin{itemize}
    \item We propose a novel progressive framework, in which a Motion Estimation (ME) module and an Event Denoising (ED) module are jointly optimized in a mutually reinforced manner, leading to more accurate motion segmentation results.
    
    \item We devise a ME module and an ED module to estimate the temporal correlation and perceive noise distribution respectively and leverage them as constraints to facilitate the mutual learning process.
    
    \item Experimental results on synthetic and real-world event sequences demonstrate the superiority of our method over the existing methods.
\end{itemize}
\section{Related Work}
In this section, we briefly review the most related event-based motion segmentation and denoising methods.

\textbf{Motion Segmentation.\quad} In recent years, event-based motion segmentation has received great attention, and a number of methods have been proposed \cite{vasco2017independent,SOFAS,stoffregen2019event, mitrokhin2020learning,kepple2020jointly,glover2017robust,mishra2017saccade,zhou2021event}. Typically, G. Guillermo \textit{et al}. \cite{gallego2018unifying} proposes a contrast maximization framework to conduct motion segmentation by maximizing the contrast of warped event images. T. Stoffregen \textit{et al}. \cite{stoffregen2019event} further extends the contrast maximization mechanism to multi-object motion estimation problem and outputs segmentation probability. Specifically, the sharpness of compensated event image measured by contrast is maximized for each motion group and used as a reference to update the segmentation probability. In \cite{mitrokhin2020learning}, a representation of events as a 3D graph is proposed to learn scene motion segmentation by a moving camera. However, since the above methods do not take background noise in event stream into account, the performance degrades seriously in the case of insufficient or changing illumination.

\begin{figure*}
    \begin{center}
        \includegraphics[width=0.9\textwidth]{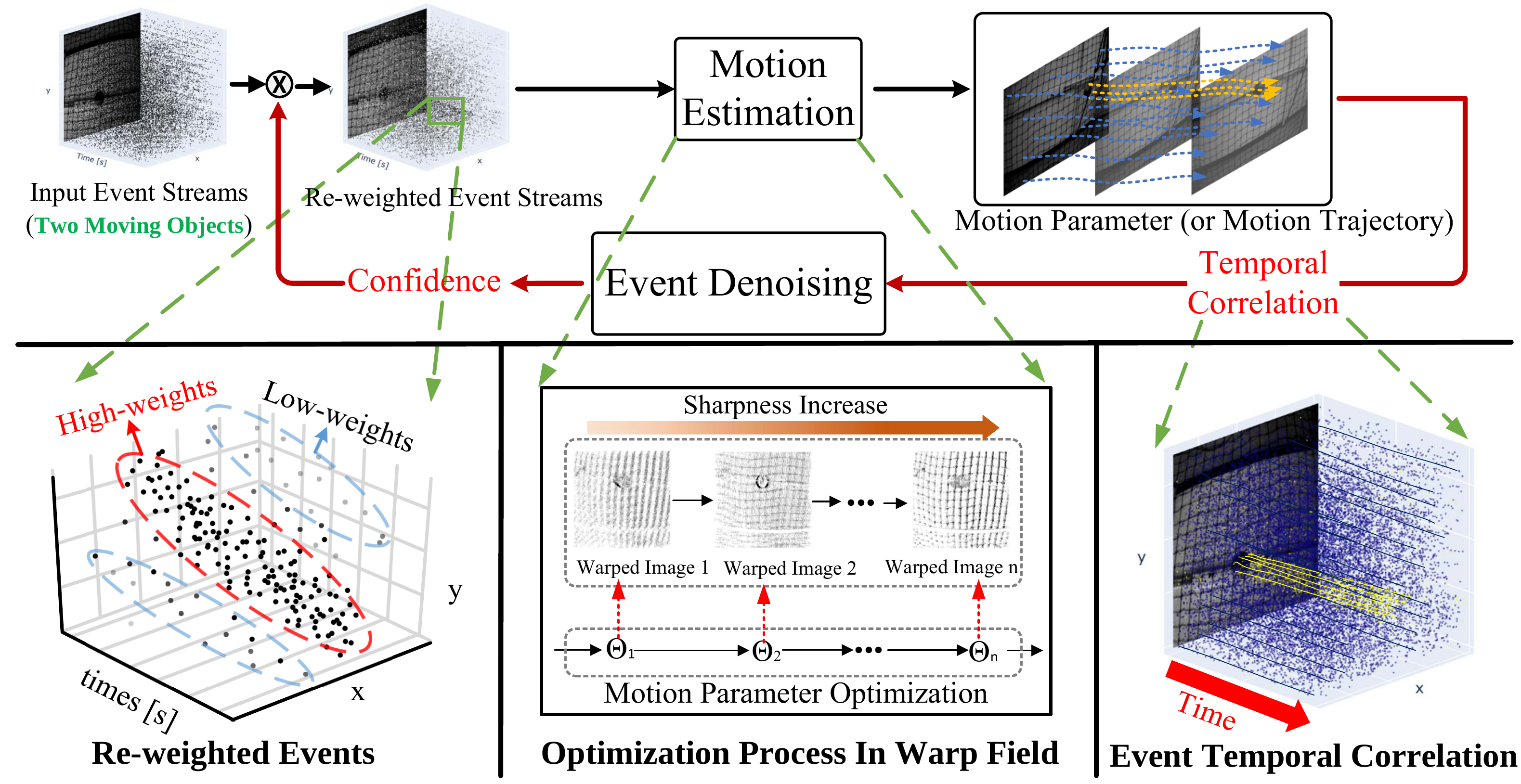}
    \end{center}

    \caption{Overview of our proposed progressive motion segmentation framework. Given an event sequence, the motion parameters $\theta$ are optimized in a warp field space under the constraint of maximum sharpness criterion. The motion parameter represents the trajectory of moving objects, which is utilized to calculate the event temporal correlation. Then, the temporal correlation is transmitted to event denoising module to guide the event re-weighting. The motion estimation and event denoising are progressively iterated until obtaining a stable motion segmentation result.}
    \label{fig:cycle}
\end{figure*}
\textbf{Event Denoising.\quad} Due to the influence of dark current and junction leakage current, the output of the event camera will inevitably contain enormous noise. To improve the quality of event stream, a lot of event denoising methods are proposed \cite{khodamoradi2018n,liu2015design,delbruck2008frame,feng2020event,baldwin2020event,wu2020denoising,wu2020probabilistic}. H Liu \emph{et al.} first proposes a spatiotemporal correlation filter to remove the BA noise by utilizing the distribution difference between BA noise and real activity events.To improve the effeciency of event denoising, \cite{khodamoradi2018n} proposes a O(N) event denoising method that can be directly incorporated into circuits. In \cite{feng2020event} the influence of hot pixels is further explored for better denoising results. R Baldwin \emph{et al.} \cite{baldwin2020event} proposes a CNN-based event denoising method by calculating the likelihood of generating an event at each pixel within a short time window. However, the above event denoising methods mainly utilize temporal correlation within the local spatial neighborhood and failed to capture long-time event features, which is essential for motion segmentation.
\section{Proposed Method}
\subsection{Overview of Progressive Framework}
Given a packet of noisy events, our goal is to segment real activity events into independently-moving objects while suppressing the influence of noise. To this end, we propose a novel progressive optimization framework by optimizing the motion estimation and event denoising in a mutually reinforced manner. As shown in Fig.\ref{fig:cycle}, our proposed progressive framework consists of two modules: Motion Estimation (ME) module and Event Denoising (ED) module.

Taking event stream as input, the ME module first conducts adaptive clustering by warping the events to a reference time and maximizing event alignment, \emph{i.e.}, maximizing the sharpness of motion-compensated images of warped events \cite{gallego2018unifying,stoffregen2019event}. The motion parameters are estimated based on the clustering results. After that, the temporal correlation of event stream is calculated by utilizing the estimated motion parameters and transmitted to ED module to guide event denoising. Taking the temporal correlation map as input, ED module first maps the correlation value to event confidence, ranging from 0 to 1. The confidence value represents the probability of an event belonging to real activity events, and the greater the confidence value, the greater the possibility of being a real activity event. Then, the confidence value is multiplied to the event, which is transmitted to ME module to update the motion estimation. Each of the preceding two processes benefits from the gradual improvement results in the other. We progressively perform the two processes until obtaining stable results.

\subsection{The ME Module}
The DVS can record logarithmic intensity changes with millisecond resolution:  
\begin{equation}
\Delta L({x_k},{t_k}) = L({x_k},{t_k}) - L({x_k},{t_k} - \Delta {t_k}) = {b_k}C.
\end{equation}
where $L(\bm{x},t)=logI(\bm{x},t)$. $\bm{x}$, $C$, $t_k$ and polarity $b_k$ represent the pixel location, contrast threshold, timestamp of the event and direction of intensity changes, respectively. The DVS will output a series of ``events'' when the change in intensity at pixel $x_k$ reaches a threshold $C$:

\begin{equation}
E=\{e_k=(\bm{x}_k,t_k,b_k)|k=1,2...N_e\}.
\end{equation}

Given a series of events and a reference time (\emph{e.g.}, $t_{ref} = 0$), ME module divides the input events into several segments by adaptive clustering in a motion compensating warp field to build several Weighted Images of Warped Events (WIWE):

\begin{equation}\label{WIWE}
\mathrm{WIWE}(\bm{x})=\sum\limits_{k = 1}^{{N_e}} {{c_{kj}}} {p_{kj}}\delta (x - {{x'}_{kj}}).
\end{equation}
Here ${c_{kj}} \in [0,1]$ represents the confidence of event belonging to real activity event, which will be explained in the following section. ${{p_{kj}}}=P({e_k} \in {\ell _j})$ represents the probability of the k-th event belonging to the j-th cluster (or motion), $\bm{x}_{kj}'=\bm{\mathrm{W}}(\bm{x}_k,t_k;\bm{\theta}_j)$ is the warped event location under the constraint of the j-th warping parameter $\bm{\theta}_j$, $\delta$ is Dirac delta. Different from the IWE in \cite{stoffregen2019event}, our proposed WIWE can better suppress the influence of BA noise. After warping, the events can be represented as:
\begin{equation}
e_k=(\bm{x}_k,t_k,b_k)\mapsto e_k'=(\bm{x}_k',t_{ref},b_k).
\end{equation}

\begin{figure*}
    \begin{center}
        \includegraphics[width=0.85\textwidth]{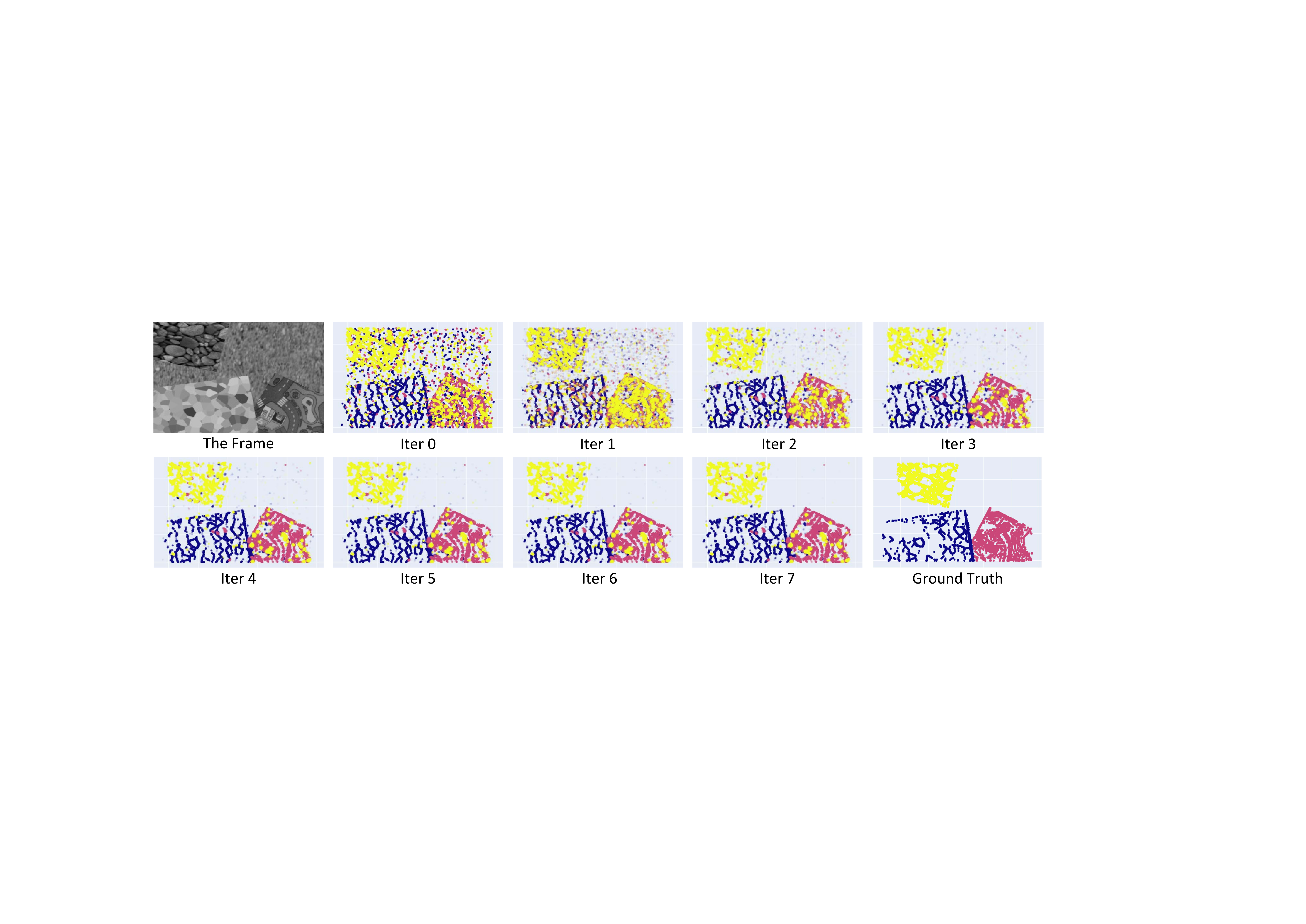}   
    \end{center}
    \caption{The motion segmentation results output by the models with different iterations. As the number of iterations increases, the influence of noise on motion segmentation is gradually suppressed.}
    \label{fig:progressive}
\end{figure*}

\begin{figure}
    \centering
    \includegraphics[width=0.35\textwidth]{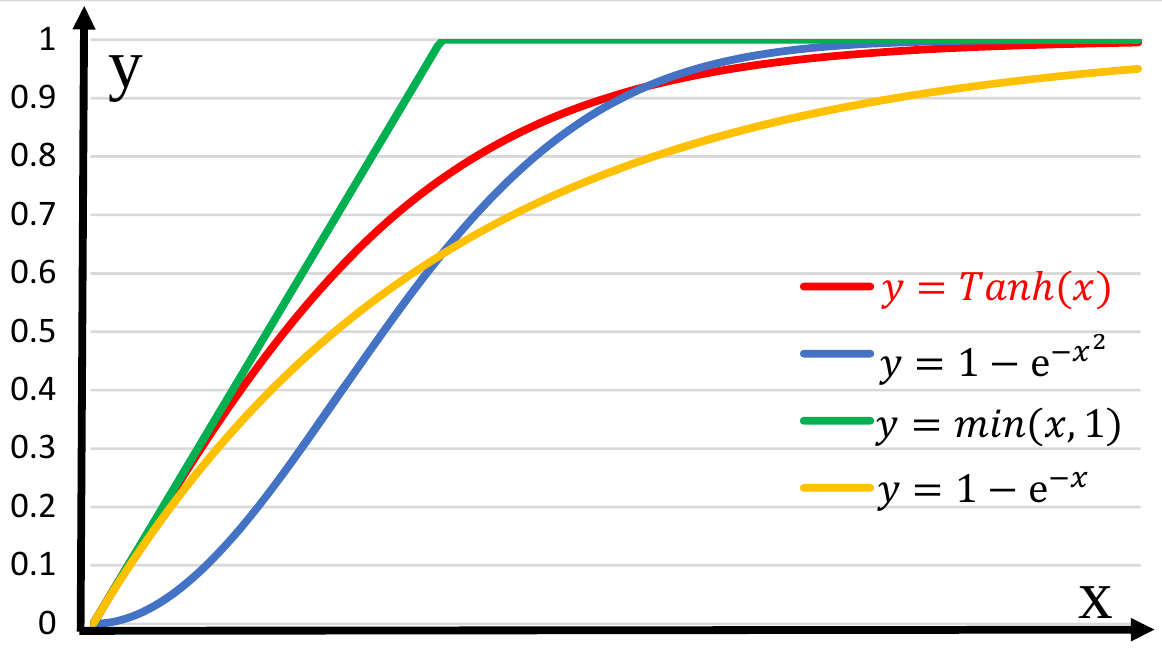}
    \caption{The confidence mapping functions.}
    \label{fig:mappingfunction}
\end{figure}

\textbf{\R{Cost} Function.\quad} To measure the event alignment within the $j$-th cluster, we propose a novel sharpness metric and use it as \R{cost} function:
\begin{equation}\label{Sharpness}
\mathrm{Sharp}(\mathrm{WIWE}_j) = \frac{\sum_k \mathrm{Var}_{jk}\times |\mathrm{IEC}_{jk}|}{\sum_k |\mathrm{IEC}_{jk}|},
\end{equation}
where $\mathrm{Var}_{jk}$ represents the local variance of \R{the $j$th} image of warped events at pixel $k$, which is defined as 
\begin{equation}
\mathrm{Var}_{ik}=\frac{1}{|\omega|^2}\sum_{j\in\omega_k} (\mathrm{WIWE}_{ij}-\mu_{ik})^2,
\end{equation}
where $\omega$ is the window size and $\mu_{ik}$ is the local mean around pixel $k$.
IEC represents the image of events correlation, which is defined as:
\begin{equation}
\mathrm{IEC}(\bm{x})=\sum_{k=1}^N b_k\delta(\bm{x-x_k'})/(t_1-t_0),
\end{equation}
where N represents the number of events between $t_0$ and $t_1$, the event starting time and the event ending time.

The IEC is derived from the optical flow constraint equation \cite{horn1981determining} and working principle of DVS:
\begin{equation}\label{flow}
\frac{\partial L}{\partial t}(\bm{x}) = -\nabla L(\bm{x}) \cdot v,
\end{equation}
\begin{equation}\label{Priciple}
\frac{\partial L}{\partial t}(\bm{x}) \approx C \frac{\sum_{k=1}^{N}b_k\delta(\bm{x}-\bm{x}_k)}{\Delta t},
\end{equation}
where $v$ represents the optical flow, $N$ represents the number of events within $\Delta t$. Based on Eq.~\eqref{flow} and Eq.~\eqref{Priciple}, we can get the following formulation:
\begin{equation}\label{MM}
    \int_{t_0}^{t_1}\nabla I \cdot v dt = -C\times \sum_{k=1}^{N}b_k \delta(\bm{x}-\bm{x}_k')/(t_1-t_0).
\end{equation}
The right side of Eq.\eqref{MM} is denoted as temporal correlation (\emph{i.e.}, IEC), which represents the intensity of projected gradient of scene. Since real activity events are caused by moving edges \cite{tpami2020survey}, they are highly correlated in time.
Different from real activity events, the noise is randomly distributed, thus it has low temporal correlation.

Under the constrained of Eq. \eqref{Sharpness}, we maximize the sum of sharpness of all clusters to find the optimal motion parameters and event confidence. After that, we update the motion segmentation results as in \cite{stoffregen2019event}:
\begin{equation}
    p_{kj}=\frac{\mathrm{WIWE}_j(\bm{x}_k')}{\sum_{i=1}^{N_l}\mathrm{WIWE}_i(\bm{x}_k')},
\end{equation}
where $N_l$ is the number of candidate motions.

\begin{figure*}
    \centering
    \begin{center}
        \includegraphics[width=0.9\textwidth]{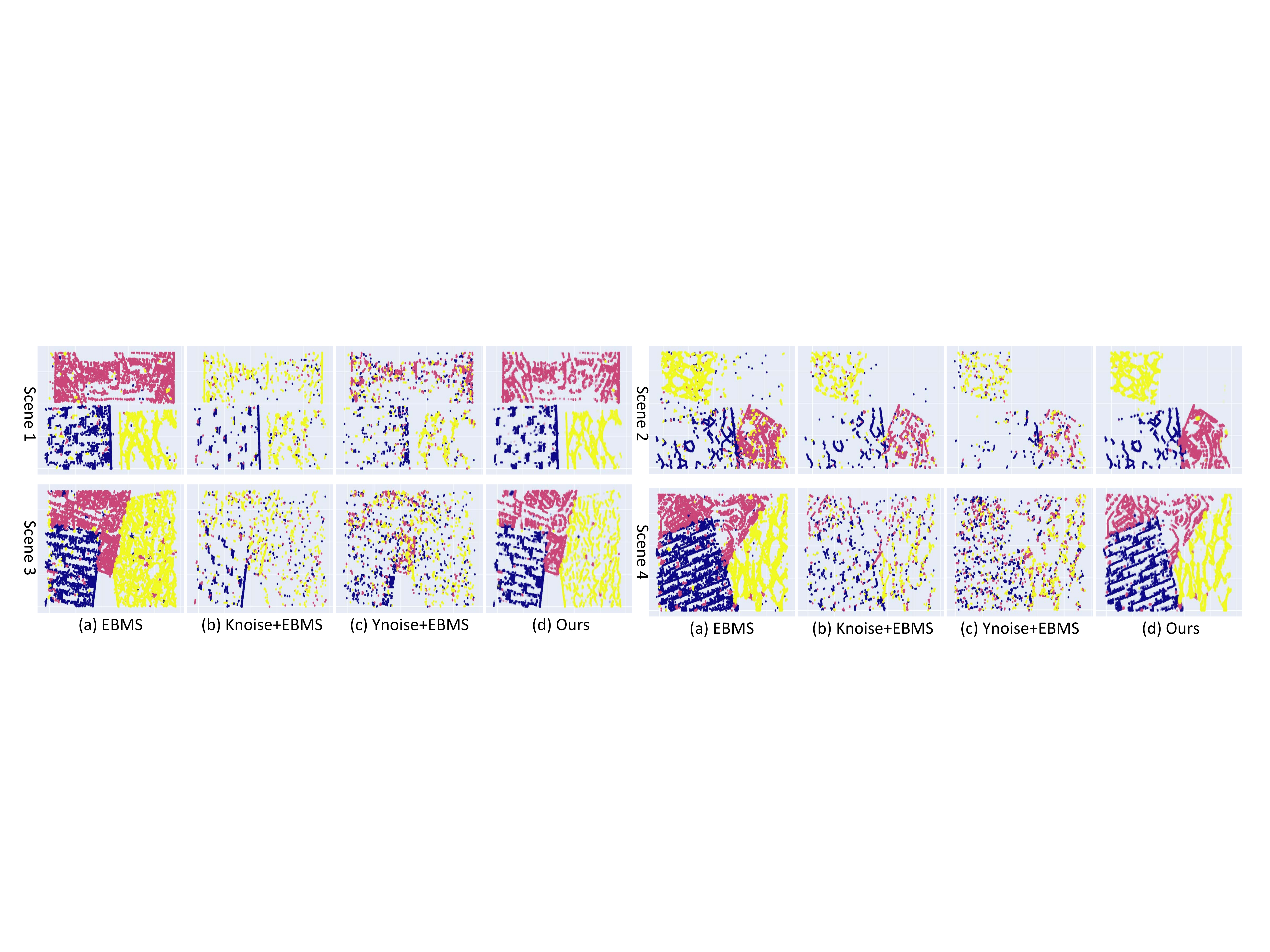}
    \end{center}

    \caption{Comparison of motion segmentation results on four synthetic scenes. Our proposed method maintain the distribution of original real activity events while suppressing the influence of noise, thereby obtaining better motion segmentation results.}
    \label{fig:env}
\end{figure*}

\begin{figure}
    \centering
    \includegraphics[width=0.5\textwidth]{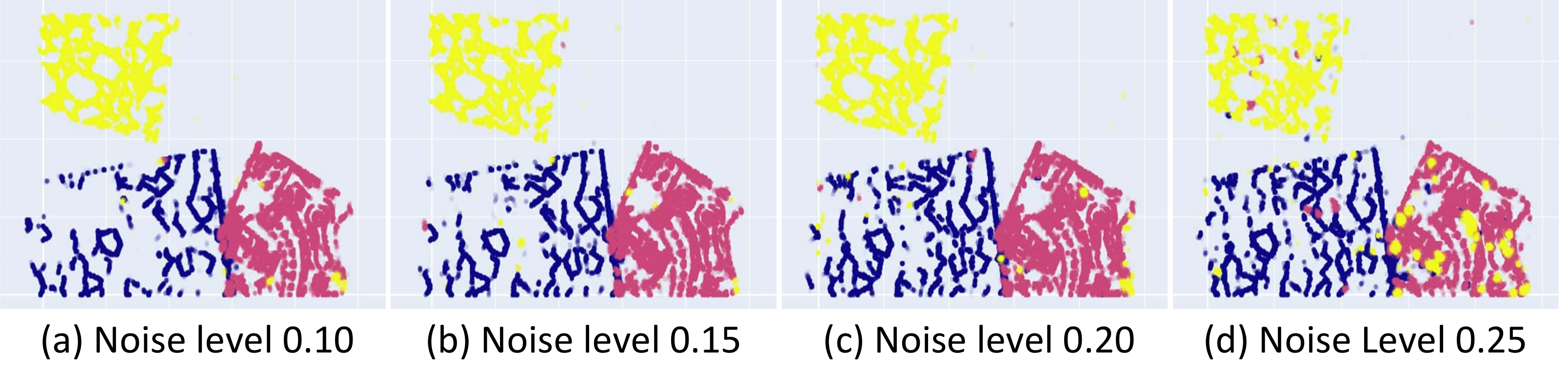}
    \caption{Motion segmentation results of our proposed method on different noise levels conditions.}
    \label{fig:robustness_to_noise}
\end{figure}

\subsection{The ED Module}
\R{This module is devised to assign a weight for each event by utilizing temporal correlation (IEC) output by Motion Estimation (ME) module. Firstly, given IEC as input, ED module takes the absolute value of IEC to eliminate the influence of direction; Secondly, the correlation map is normalized to the range of [0,1]. Events with smaller weights are more likely to belong to noise. Finally, the weight is multiplied by the event and transferred to the ME module to suppress the influence of noise in motion estimation.}

\R{Specifically,} the ED Module is devised to assign confidence to each event, representing the probability of an event belonging to real activity events. Theoretically, if an event belongs to BA noise, its confidence should be close to zero. To achieve this, we first take the absolute value of IEC to discard the influence of the direction of projection:
\begin{equation}
\mathrm{EC}_{ij}=|\mathrm{IEC}_j(\bm{x_i'})|.
\end{equation}
Then, the correlation is mapped to the range of [0,1]. In this paper, we compare four kinds of mapping functions, as shown in Fig.\ref{fig:mappingfunction}. We find the Tanh function performs best: 
\begin{equation}
    c_{ij}=tanh(\lambda \mathrm{EC}_{ij}).
\end{equation}
Here $\lambda$ is the normalizing factor that controls the denoising level, which is set as the reciprocal of the mean of $EC$:
\begin{equation}
\lambda  = \frac{1}{\mathrm{Mean}(\mathrm{EC})}.
\end{equation}
After that, we can obtain a $N\times N_l$ confidence map, which is used to re-weight the events as in Eq. \eqref{WIWE}. Finally, the re-weighted events are used as the input of ME module to update motion segmentation results.

\section{Experiments}
To demonstrate the superiority of our progressive framework for motion segmentation, both quantitative and qualitative evaluations on synthetic and real-world event streams are conducted. We compare our model with SOTA methods, including event based motion segmentation method, EBMS \cite{stoffregen2019event} and event denoising methods, Knoise \cite{khodamoradi2018n} and Ynoise \cite{feng2020event}. For each event denoising method, we first filter out the noise and then conduct segmentation using EBMS.

\subsection{Event Dataset and Evaluation Metric}
\textbf{Synthetic DVS Data.\quad} The synthetic DVS data are generated using ESIM \cite{rebecq2018esim} on simulated scenes, created with blender \cite{blender}. Each simulated scene contains three objects with different depths, ranging from 0.5 to 3. To simulate the distribution of objects in the real environment, we set occlusions between different objects. The camera moves parallel to objects with a speed of 0.4/s. The contrast threshold is 0.5. Then, we add Gaussian noise as in \cite{patrick2008128x}. To thoroughly verify the robustness of the proposed method, we set five kinds of noise levels ($n \in \{ 0.05,\;0.10,\;0.15,\;0.20,\;0.25\}$) for each simulated scene. In this paper, we create four simulated scenes, and thus obtaining a total of 20 synthetic event sequences. 

\textbf{Real-world DVS Data.\quad} The Extreme Event Dataset (EED) \cite{mitrokhin2018event} is a real-world event segmentation benchmark, which contains both event sequences with light noise level, \emph{i.e.}, \textit{``what is background"}, and event sequences with heavy noise level, \emph{i.e.}, \textit{``light variations"}. \R{The EV-IMO dataset \cite{mitrokhin2019ev} is an object-level motion segmentation dataset aiming at distinguishing each moving object and estimate its motion parameters.} We use \R{both of them} to evaluate the effectiveness of our proposed method to real-world noisy event sequences.

\textbf{Evaluation Metric.\quad}
For synthetic event sequences, we adopt the widely used Intersection over Union (IoU) \cite{jaccard1912distribution} metric for quantitative evaluation. For sub-set events with the same motion trajectory, we count the number of True Positive (TP), False Positive (FP), and False Negative (FN) events, respectively, and calculate the IoU as:

\begin{equation}
\mathrm{IoU} = \frac{\mathrm{TP}}{\mathrm{TP} + \mathrm{FP} + \mathrm{FN}} \times 100\%.
\end{equation}
The higher the IoU value, the higher the motion segmentation accuracy. For the scenes with multi-moving objects, we calculate the IoU of each object respectively and use the mean of IoU, denoted as MIoU, to represent the motion segmentation accuracy of the test scene. For the EED dataset, we use the provided timestamped bounding boxes to calculate the percentage of segmented events lying inside the bounding box after the warping process, and use it to represent the motion segmentation accuracy. \R{We also present the object segmentation success (OSS) rate results proposed by \cite{mitrokhin2018event}.}

\subsection{Ablation Study}
\textbf{Number of Iterations.\quad}To examine the improvements induced by diﬀerent iterations, we experimentally compare motion segmentation result with diﬀerent iterations on synthetic events ($n=0.25$). And the results are presented in Fig.\ref{fig:progressive}. Compared with only performing motion estimation, \emph{i.e.}, iteration is 0, the introduction of event denoising can eﬀectively improve the accuracy for motion segmentation. As the number of iterations increases, the effect of noise is gradually suppressed and reaches stability in the seventh iteration. This demonstrates that the progressive iteration mechanism between motion estimation and event denoising can improve each other. In the following sections, the model with seven iterations is used as our default model.


\textbf{\R{Cost} Function.\quad} To demonstrate the effectiveness of our proposed loss function, we compare it with several widely used loss functions proposed in \cite{gallego2019focus}. We conduct the experiment on simulated scenes with noise level ($n=0.25$), and MIoU results are reported in Table.\ref{tab:loss}. It is clear that our proposed loss function outperforms other loss functions. That is due to our loss function can better suppress the interference of noise by utilizing structural information.

\begin{figure*}
    \begin{center}
        \includegraphics[width=0.9\textwidth]{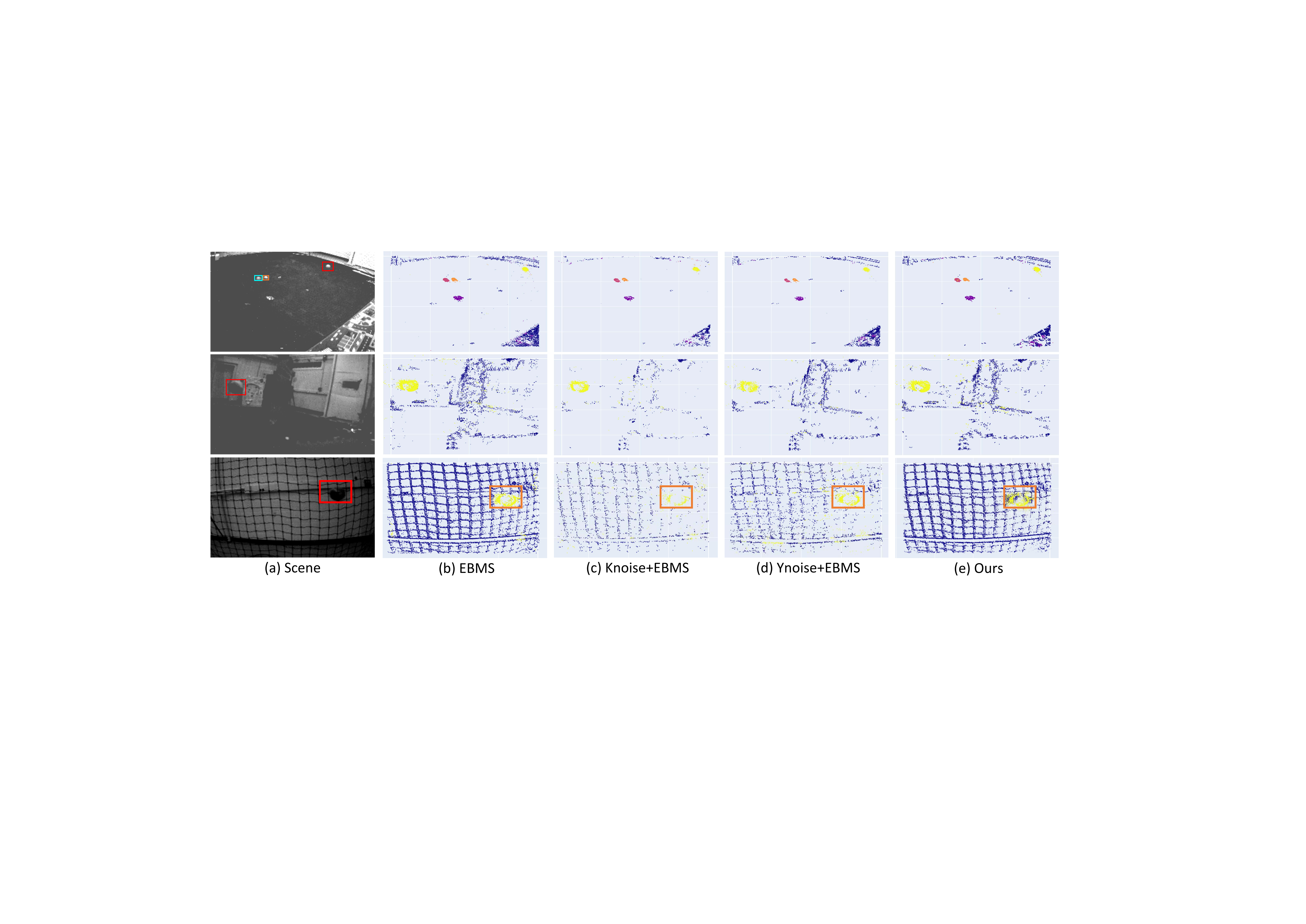}
    \end{center}
    \caption{Segmentation results on real-world event sequences with light noise level. Orange boxes highlight segmentation difference between direct segmentation and our method.}
    \label{fig:real-lightnoise}
\end{figure*}

\begin{table}[]
\small
\caption{Comparison results with existing loss functions for DVS (in \%). VI, MIG and MIH represent Image Variance, Magnitude of Image Gradient and Magnitude of Image Hessian \cite{gallego2019focus}, respectively.}
    \centering
    \begin{tabular}{|c|c|c|c|}
    \hline 
        VI & MIG & MIH & {\color[HTML]{FF0000} \textbf{Ours}} \\
        \hline
        73.03 & 59.16 & 72.52 & {\color[HTML]{FF0000} \textbf{79.52}} \\
    \hline
    \end{tabular}
    \label{tab:loss}
\end{table}

\begin{table}[]
\small
\centering
\caption{Comparison results with different methods on five kinds of noise conditions (in $\%$).}
\begin{tabular}{|c|ccccc|}
\hline
NoiseLevel  & 0.05    & 0.10    & 0.15    & 0.20    & 0.25     \\ \hline
Knoise+EBMS & 89.15   & 86.51   &  81.92  &  78.93   & 38.18  \\
Ynoise+EBMS & 84.10   & 81.66  & 77.89  &   71.09   &  54.11    \\
EBMS        & 90.82   & 88.97  & 87.15  &   81.53  & 60.61  \\ \hline
Ours        & {\color[HTML]{FF0000} \textbf{94.63}} & {\color[HTML]{FF0000} \textbf{93.71}} & {\color[HTML]{FF0000} \textbf{92.95}} & {\color[HTML]{FF0000} \textbf{89.99}} & {\color[HTML]{FF0000} \textbf{79.52}} \\ \hline
\end{tabular}
\label{tab:table1}%
\end{table}


\subsection{Comparisons on Synthetic DVS Data}
\textbf{Quantitative Results.\quad} We report the MIoU results of different methods on five kinds of noise levels, as shown in Table.\ref{tab:table1}. It is clear that our method achieves the best overall performances on all noise levels. On low noise level conditions (\emph{e.g.}, $n$ = 0.05, 0.10 ), although the EBMS method can achieve good motion segmentation results, the accuracy is still lower than our method. The reason is that EBMS doesn't consider the effect of noise. Besides, although Knoise and Ynoise denoising algorithms can filter out noise, they will also cause the loss of valid events, resulting in incomplete motion segmentation. Unlike these methods, our method can adaptively perceive the noise distribution without damaging the distribution of real activity events and thus obtaining better motion segmentation results. Furthermore, as the noise level increase, the performance of existing methods drops significantly, but our method can still hold ideal motion segmentation performance on heavy noise level conditions.

\begin{table*}[]
\small
\caption{\R{MIoU} results with different methods on EED dataset (in $\%$). FMD, LV, WIB, OC and MO represnet the \textit{``Fast moving drone"}, \textit{``Lighting variations"}, \textit{``What is background?"}, \textit{``Occluded sequence"}, \textit{``Multiple objects scene"}, respectively.}
\centering
\begin{tabular}{|c|cc|ccc|}
\hline
 & \multicolumn{2}{l|}{Heavy Noise Level (HNL)} & \multicolumn{3}{l|}{Light Noise Level (LNL)} \\ 
\cline{2-6}
\multirow{-2}{*}{EED Sequence Name (ESN)}  & FMD  & LV  & WIB & OC  & MO \\ \hline
Knoise+EBMS  & 47.22 & 31.08 & 69.06 & 82.41 & 42.74 \\
Ynoise+EBMS  & 25.63 & 34.49 & 63.99 & {\color[HTML]{FF0000} \textbf{86.12}} & 41.32 \\
EBMS & 27.17 & 30.95  & 68.43  & 69.66 & 41.63 \\ \hline
{\color[HTML]{FF0000} \textbf{Ours}} & {\color[HTML]{FF0000} \textbf{53.92}} & {\color[HTML]{FF0000} \textbf{48.33}} & {\color[HTML]{FF0000} \textbf{93.33}} & 83.23 & {\color[HTML]{FF0000} \textbf{51.11}} \\ \hline
\end{tabular}
\label{tab:table2}%
\end{table*}

\begin{table}[]
\small
\caption{\R{OSS results with different methods on EED dataset. EMODT refers to the method in \cite{mitrokhin2018event}.}}
\centering
\begin{tabular}{|c|cc|ccc|}
\hline
                                     & \multicolumn{2}{c|}{HNL}                                                       & \multicolumn{3}{c|}{LNL}                                                                                                \\ \cline{2-6} 
\multirow{-2}{*}{ESN}                & FMD                                    & LV                                    & WIB                                    & OC                                    & MO                                     \\ \hline
EMODT                            & 92.78                                  & 84.52                                 & 89.21                                  & 90.83                                 & 87.32                                  \\
EBMS                                 & 96.30                                  & 80.51                                 & {\color[HTML]{FF0000} \textbf{100.00}} & {\color[HTML]{FF0000} \textbf{92.31}} & 96.77  \\ \hline
{\color[HTML]{FF0000} \textbf{Ours}} & {\color[HTML]{FF0000} \textbf{100.00}} & {\color[HTML]{FF0000} \textbf{90.90}} & {\color[HTML]{FF0000} \textbf{100.00}} & {\color[HTML]{FF0000} \textbf{92.31}} & {\color[HTML]{FF0000} \textbf{100.00}} \\ \hline
\end{tabular}
\label{tab:OSS}
\end{table}

\begin{figure*}
    \begin{center}
        \includegraphics[width=0.9\textwidth]{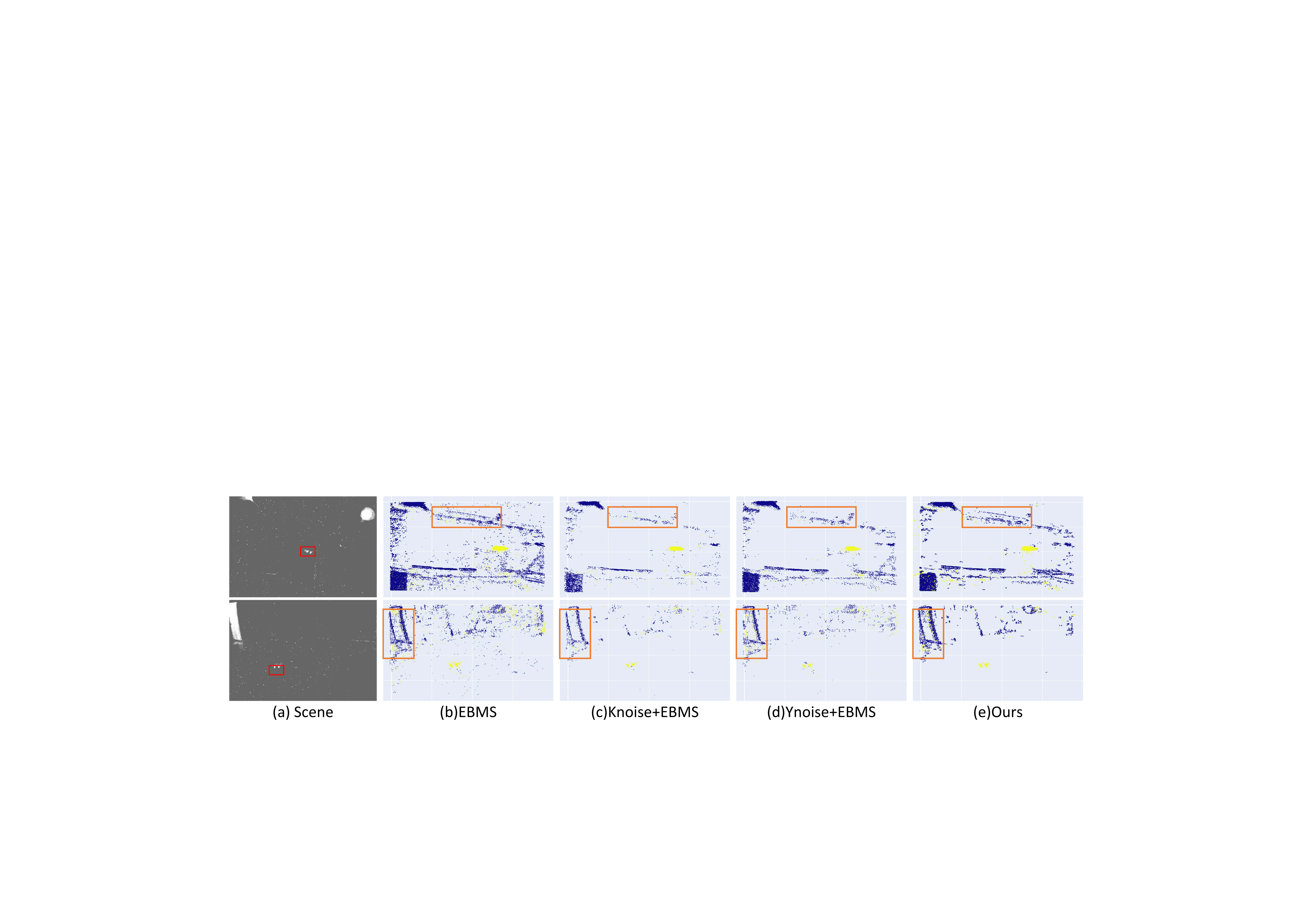}
    \end{center}
    \caption{Segmentation results on real-world event sequences with heavy noise level. Orange boxes highlight segmentation difference between direct segmentation and our method.}
    \label{fig:real-heavynoise}
\end{figure*}

\textbf{Qualitative Visual Results.\quad}
We first present the motion segmentation results on four simulated scenes with noise level $n$=0.05, as shown in Fig.\ref{fig:env}. We can observe that our method outperforms existing methods greatly. Specifically, compared with EBMS, not only the events that lie outside valid regions are suppressed, but also the erroneously segmented events are corrected, as shown in scene 2. This is because in our framework, the progressively updated event confidence can provide more accurate event correlation. As is shown in Fig.\ref{fig:env} (b-c), large amounts of useful information are also filtered out along with event noise by Knoise or Ynoise. Compared with them, our method can perceive noise in an adaptive way and thus has better performance. 

We also present the motion segmentation results on different noise level conditions, as shown in Fig.\ref{fig:robustness_to_noise}. We can observe that as the noise level increases, our method can always effectively suppress the influence of noise on motion segmentation and obtain complete segmentation results. This demonstrates that the proposed method can accurately perceive the noise distribution and maintain robust under different noise levels.

\subsection{Comparisons on Real-world DVS Data}
\textbf{Quantitative Results.\quad}
The EED dataset \cite{mitrokhin2018event} is an event segmentation benchmark which contains both event sequences generated on normal lighting conditions, \emph{e.g.}, \textit{``multiple objects"} and \textit{``what is background"}, and event sequences generated on low-light lighting conditions, \emph{e.g.}, \textit{``fast moving drone"} and \textit{``lighting variation"}. According to the difference of light intensity, we divided event sequences in EED into two groups, \emph{i.e.}, low noise level and high noise level, and reported the results of different methods on these two groups of sequences as shown in Table.\ref{tab:table2} \R{and \ref{tab:OSS}}.

As shown in Table.\ref{tab:table2}, the accuracy of motion segmentation on heavy noise conditions is much lower than that in light noise conditions. In general, the EBMS achieves the best performance among three comparison methods. However, the accuracy of EBMS on heavy noise level conditions is still very low. For example, the accuracy of EBMS on \textit{``fast motion drone"} is only 27.17\%. This is due to the temporal correlation of real activity events is greatly destroyed. Different from the existing methods, our proposed method can effectively improve the motion segmentation accuracy. For example, our method achieves an accuracy of 53.92\% on \textit{``fast moving drone"}, which is 26.75\% higher than EBMS. Besides, our method can also improve the motion segmentation performance on light noise conditions as shown in Table.\ref{tab:table2}. This demonstrates the effectiveness of our method on real-world event sequences, which is consistent with the conclusion on synthetic event data. \R{Segmentation results as OSS rates are shown in Table.\ref{tab:OSS}. As can be seen, our method outperforms all existing methods especially on high noise level conditions, and can even achieve 100\% object detection rate on the FMD, WIB and MO sequences.} 

\begin{figure}
    \begin{center}
        \includegraphics[width=0.443\textwidth]{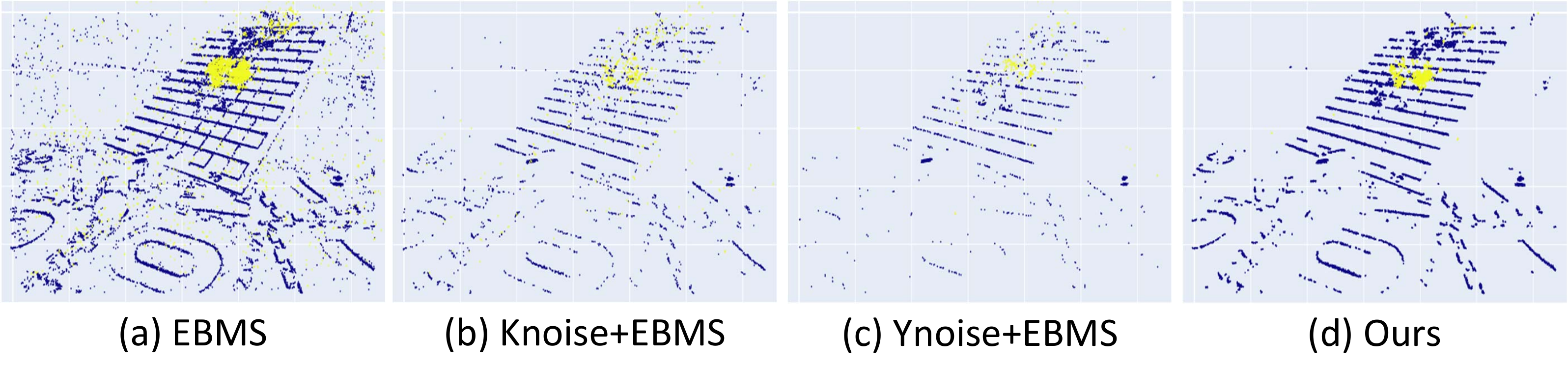}
    \end{center}
    \caption{\R{Segmentation results on EVIMO event sequence with heavy noise level. In this sequence a toy car is sliding down from a slope and the camera is randomly moving.}}
    \label{fig:evimo}
\end{figure}

\textbf{Qualitative Visual Results.\quad} Then, we visualize motion segmentation results on light and heavy noise conditions \R{in EED dataset} as shown in Fig.\ref{fig:real-lightnoise} and  Fig.\ref{fig:real-heavynoise}. It is clear that our method can effectively suppress the influence of noise as well as hold the original structures of moving objects. Our method can not only suppress the influence of noise but also maintain the original effective structures, which demonstrates the effectiveness of our method for real-world event sequences. \R{Besides, we also provided the segmentation results of EVIMO dataset \cite{mitrokhin2019ev} where different colors represent different motions as in Fig.\ref{fig:evimo}. As can be seen, our method preserves most useful structure information while at the same time noisy events are eliminated.}


\section{Conclusion}

In this paper, we present a novel progressive framework to improve the quality of motion segmentation under noisy environments in a mutually reinforced manner. The proposed framework consists of a Motion Estimation (ME) module and an Event Denoising (ED) module which are jointly optimized by maximizing sharpness metric and each module can benefit from the gradually improved results in the other. The comprehensive evaluations on both synthetic and real event sequences demonstrate that our proposed method achieves superior performance over the SOTA methods.

\section{Acknowledgements}
\R{This work was supported by the National Key R\&D Program of China under Grand 2020AAA0105700, National Natural Science Foundation of China (NSFC) under Grants U19B2038, 61872327, the University Synergy Innovation Program of Anhui Province under Grants GXXT-2019-025, the key scientific technological innovation research project by Ministry of Education and Major Special Science and Technology Project of Anhui (No. 012223665049).}

\newpage 
\bibliography{aaai22}
\end{document}